\documentclass[runningheads]{llncs}
\usepackage{amsmath,amsfonts,amssymb}

\usepackage{multirow}
\usepackage{xcolor}
\usepackage[T1]{fontenc}
\usepackage{graphicx}
\begin{document}
\title{Enhanced Neuromorphic Semantic Segmentation Latency through Stream Event}
\titlerunning{Neuromorphic Semantic Segmentation Leveraging Stream Event}
%
\author{Dalia Hareb\inst{1,2}\orcidID{0009-0004-3022-8521} \and 
Jean Martinet\inst{1}\orcidID{0000-0001-8821-5556} \and
Benoit Miramond\inst{2}\orcidID{0000-0002-1229-7046}}

\authorrunning{D.~Hareb, J.~Martinet, B.~Miramond}
%
\institute{Université Côte d’Azur, CNRS, I3S, France \and
LEAT, Université Côte d'Azur, Sophia Antipolis, France \\
\email{\{firstname.secondname\}@univ-cotedazur.fr}}
\maketitle              
\begin{abstract}
Achieving optimal semantic segmentation with frame-based vision sensors poses significant challenges for real-time systems like UAVs and self-driving cars, which require rapid and precise processing. Traditional frame-based methods often struggle to balance latency, accuracy, and energy efficiency. To address these challenges, we leverage event streams from event-based cameras—bio-inspired sensors that trigger events in response to changes in the scene. Specifically, we analyze the number of events triggered between successive frames, with a high number indicating significant changes and a low number indicating minimal changes. We exploit this event information to solve the semantic segmentation task by employing a Spiking Neural Network (SNN), a bio-inspired computing paradigm known for its low energy consumption. Our experiments on the DSEC dataset show that our approach significantly reduces latency with only a limited drop in accuracy. Additionally, by using SNNs, we achieve low power consumption, making our method suitable for energy-constrained real-time applications. To the best of our knowledge, our approach is the first to effectively balance reduced latency, minimal accuracy loss, and energy efficiency using events stream to enhance semantic segmentation in dynamic and resource-limited environments.
\end{abstract}
\section{Introduction}
Event-based cameras have emerged as innovative instruments in computer vision, introducing novel methods for visual perception. 
They boast exceptional temporal resolution,
capturing data in the order of microseconds, and offering a dynamic range exceeding $120db$, far surpassing conventional standards having $60db$. Furthermore, these cameras demonstrate minimal latency, with a delay as low as $10ms$, and are designed for efficient power consumption, capped at $100mW$. In contrast to traditional frame cameras, which capture video as a sequence of dense frames at a fixed rate, event cameras operate asynchronously, detecting per-pixel brightness changes and encoding them into a continuous stream of events denoted by $<x, y, p, t>$, representing spatial coordinates $(x, y)$, polarity $(p)$, and time $(t)$. These events facilitate the determination of both the spatial extent and intensity of changes within the scene. For example, a high density of events triggered across specific pixels within a given time window $\delta t$ denotes a substantial influx of new information. Conversely, a sparse occurrence of events indicates minimal changes occurring within the corresponding pixels. This observation underlies the exploration in this paper, leveraging its significance to streamline computing, reduce power consumption, and minimize latency by focusing solely on regions of images with significant changes.

\noindent With the continual aim of decreasing energy consumption, our focus shifted from standard ANNs to SNNs \cite{maass1997networks} for image processing. These neural networks diverge from ANNs by employing an asynchronous sparse event-driven computation paradigm for processing data in a spatio-temporal manner. Within SNNs, spikes serve as information units, and membrane potential represents the neuron's state. Initially set to a rest value $V_0$, typically $0V$, the membrane potential increases as it accumulates the weights of synapses activated by pre-synaptic neuron spikes. This accumulation halts once the membrane potential surpasses a predefined threshold. Upon reaching this threshold, the membrane potential undergoes either a hard reset to $V_0$ or a soft reset, where the threshold value is subtracted from it while the neuron initiates a spike. Alternatively, suppose the membrane potential fails to reach the threshold. In that case, its value decreases over time with a decay factor $\tau$ for Leaky Integrate-and-Fire (LIF) neurons \cite{nahmias2013leaky} or remains constant for Integrate-and-Fire (IF) neurons. Hence, by leveraging binary matrices that exclusively require addition operations, rather than both addition and multiplication and with energy consumption only upon spike activation, SNNs achieve substantial energy efficiency gains and reduced latency. This assertion has been proved by studies conducted by Khacef et al. \cite{khacef2018confronting}, Lemaire et al. \cite{lemaire2022analytical}, and Dampfhoffer et al. \cite{dampfhoffer2023leveraging}, who achieved comparable performance between SNNs and ANNs at lower hardware implementation costs. Consequently, this reduced power consumption addresses a critical challenge, particularly in restricting the deployment of ANNs on resource-constrained systems like self-driving cars and UAVs.

\noindent This paper introduces a novel approach to semantic segmentation, inspired by video semantic segmentation techniques based on optical flow, but adapted to leverage event-based data exclusively. The key insight is that the number of events generated between two successive frames is inversely proportional to the similarity between these frames. In scenarios where fewer events are detected, indicating high similarity, the segmentation result from the first frame is reused for the second frame. Conversely, when a significant number of events are generated, indicating substantial changes between frames, the second frame is processed using a spiking neural network, we have designed and trained on a semantic segmentation dataset. Thus, our main contributions include:

\begin{enumerate}
    \item Development of a method to exploit event-based data for semantic segmentation by dynamically adapting the processing strategy based on the event stream, bypassing the need for their structural transformations.
    \item Demonstration of reduced latency and increased throughput through computational reduction, while maintaining comparable performance.
    \item Achievement of low energy consumption by selectively activating neurons using a spiking neural network.
\end{enumerate}
The paper is structured as follows. First, a review of relevant literature in event-based semantic segmentation and video semantic segmentation is conducted in Sec. \ref{related}. Following this, a detailed description of the proposed approach is presented in Sec. \ref{method}. The approach is evaluated in Sec. \ref{exp} using the training hyperparameters outlined in Sec. \ref{impl} on the DSEC dataset described in Sec. \ref{dataset_dsec}. The results are discussed in Sec. \ref{results}. To gain a deeper understanding of the effectiveness of each component, an ablation study is conducted on DSEC in Sec. \ref{ablation}. Finally, Sec. \ref{sec_conclusion} concludes the paper and discusses future work.

\section{Related work} \label{related}
Since their emergence, event-based cameras have gained significant attention in research, particularly in robotics and computer vision domains. Numerous recent studies have leveraged these cameras for various vision tasks, including object detection, pose estimation, and semantic segmentation. As our approach addresses semantic segmentation, this paper will delve into the current state-of-the-art achievements in this area using either ANNs or SNNs. Additionally, since we aim to leverage the high time resolution of events to enhance the processing of successive images—similar to video semantic segmentation—we will review the existing works in this domain, as well.
\subsection{Event-based semantic segmentation}
The pioneering work in event-based semantic segmentation is by Alonso et al. \cite{alonso2019ev}. They introduced the Davis Driven Dataset 2017 (DDD17), a novel event-based dataset, accompanied by semantic segmentation ground truth generated using an ANN trained on Cityscape \cite{cordts2016cityscapes}. Their approach involved designing the Xception network, which accepts a $6$-channel input structure representing the mean, standard deviation, and accumulation of each polarity of events acquired over a $50ms$ interval which is $20Hz$. Subsequently, Gehrig et al. \cite{gehrig2020video} addressed the limitations of the existing dataset by augmenting the training data with synthetic events from video sources, leading to performance improvements. Later, Wang et al. \cite{wang2021dual} innovatively transformed events into images and introduced knowledge distillation (KD), aiming to transfer knowledge from labelled images to unlabeled events \cite{wang2021evdistill}. Pursuing a similar objective, Sun et al. \cite{sun2022ess} utilized unsupervised domain adaptation instead of KD. While these approaches primarily utilize ANNs for event processing, Zhang et al. \cite{zhang2023energy} and Shristi et al. \cite{das2024halsie} employed hybrid networks comprising both SNN and ANN components. Zhang et al. \cite{zhang2023energy} integrated ANNs into each block of their architecture, while Das et al. \cite{das2024halsie} fused events processed by SNN with images processed by ANN and fed the result to an ANN subnetwork. Finally, similarly to Kim et al. \cite{kim2022beyond}, we proposed fully spiking networks that process event accumulations acquired over a $50$ms interval \cite{hareb2024evsegsnn}
 
\subsection{Video semantic segmentation} 
As the demand for high accuracy and fast inference time increases, various techniques have been developed to perform video semantic segmentation more efficiently. These methods reuse features extracted by trained neural networks when there is minimal difference between consecutive video frames, thereby avoiding reprocessing every pixel in each frame using deep models. Before Transformers became widespread around $2019$, existing works such as \cite{xu2018dynamic}, \cite{li2018low}, \cite{zhu2017deep} and \cite{jain2019accel} primarily used a two-part approach. If the similarity between the current frame and a keyframe is low, the current frame is processed by a deep neural network trained in semantic segmentation.
Conversely, if the similarity is high, a warp function is applied, which involves moving the semantic segmentation result of the keyframe based on the optical flow to the current frame, followed by a bilinear interpolation to determine the exact value of each pixel. These methods have been criticized for not considering temporal information, prompting researchers \cite{wang2021end}, \cite{yang2022temporally}, and \cite{wu2022seqformer} to introduce Transformers in their architectures to capture both spatial and temporal contexts. However, since Transformers rely heavily on matrix multiplications and extensive computations, they are unsuitable for spiking neural networks. Therefore, we focused on the first approach, particularly drawing inspiration from \cite{xu2018dynamic}, since the other methods require an additional network for semantic segmentation on the current frame alongside the optical flow network, making the process slower and more computationally intensive. This method also uses dynamic video semantic segmentation, where the keyframes are changed dynamically over time rather than predefined.

\noindent Our approach addresses the semantic segmentation task using SNNs for their low power consumption. More particularly, we leverage the high time resolution from event-based cameras directly, unlike state-of-the-art methods that convert events into other formats like pseudo-frames or voxel grids \cite{gehrig2019end} to process them similarly to dense images. 
\section{Method} \label{method}

\begin{figure}[tb]
  \centering
  \includegraphics[scale=0.225]{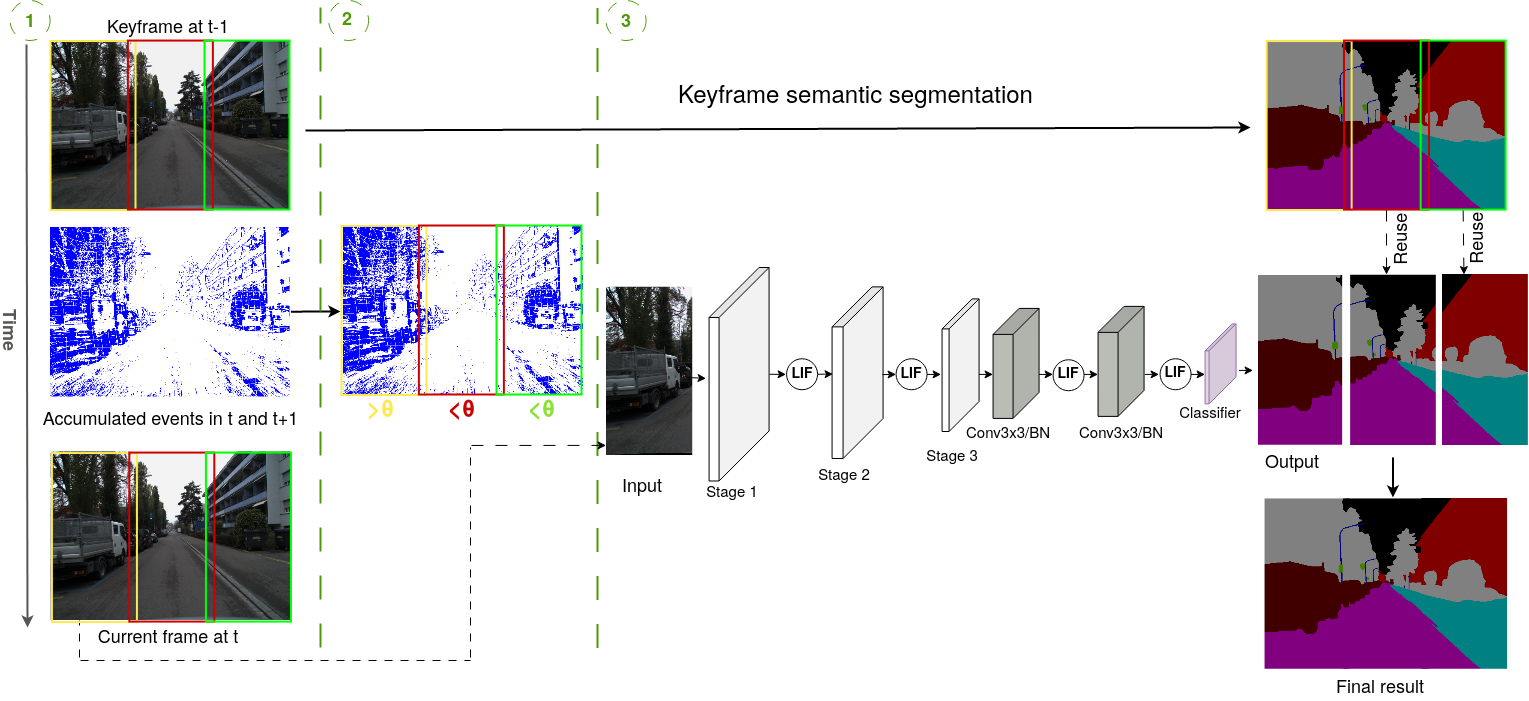}
  \caption{Network overview: The diagram illustrates the process in $3$ steps from left to right: 1) The image is divided into three overlapping regions, with the yellow, green, and red rectangles representing individual regions and their intersections representing the overlapping areas. 2) The mean number of events triggered within the time interval \([t-1, t]\) is compared against a predefined threshold $\theta$. 3) SegSNNnet processes the left region as its number of events exceeds the threshold. For the other two regions, the semantic segmentation results from the keyframe generated at time \(t-1\) are transferred to the current frame for reuse.}
  \label{fig:overview}
\end{figure}
In this section, we detail our framework approach, consisting of $3$ main steps, as illustrated in Fig. \ref{fig:overview}, along with the spiking neural network we designed for semantic segmentation, named SegSNNnet.
\begin{enumerate}
    \item \textbf{Division of the input image}: The input image is divided into \(N\) overlapping regions to account for varying changes across different parts of the image.
    \item \textbf{Event count}: The mean number of events generated in the interval $[t-1,t]$, corresponding to the time between the keyframe $t-1$ and the current frame $t$, is calculated for each region. This count is compared to a threshold to evaluate the amount of change in the region from $t-1$ to $t$.
    \item \textbf{Significant change detection}: If the event count exceeds a given threshold $\theta$, indicating a significant difference between the region belonging to the current frame and the corresponding region in the keyframe, the current frame's region undergoes semantic segmentation processing by SegSNNnet. Thus, updating the keyframe's region with the new segmentation result. Otherwise, if the event count does not exceed the threshold, implying similarity between these regions, the semantic segmentation result from the keyframe's region is directly applied to the current frame's region. Finally, the overlapping regions are merged to produce the final result.

\end{enumerate}
\noindent \textbf{SegSNNnet}: is a fully spiking architecture of $330K$ parameters, featuring a simplified ResNet18 backbone with only $3$ stages and reduced number of parameters in each block. This backbone incorporates Spike-Element-Wise (SEW) blocks, as introduced by Fang et al. \cite{fang2021deep}, which support effective residual learning in deep Spiking Neural Networks (SNNs) without the typical issues of exploding or vanishing gradients. Following the backbone, the network includes two convolutional layers with batch normalization (BN) and spiking neurons, which are succeeded by a spike accumulator that gathers spike voltages over multiple steps. This fully spiking nature is critical for deployment on neuromorphic hardware like Loihi \cite{davies2018loihi}, SpiNNaker \cite{furber2014spinnaker}, or SPLEAT \cite{Spleat_application2020,abderrahmane2022spleat}, which are tailored for SNNs on ASIC and FPGA platforms.

\noindent In the forward pass, the activation function outputs either $0$ or $1$. Consequently, during backpropagation, parameter updates involve subtracting the product of the learning rate and the gradient. However, since the derivative of the activation function is either $0$ or undefined, this can lead to weight updates being either $0$ or infinite. To overcome this issue, surrogate gradients are employed, replacing the original activation function with a substitute \cite{neftci2019surrogate}. Drawing from the approach of Guo et al. \cite{guo2023joint}, we use the rectangular function \cite{wu2019direct}, expressed as:
\begin{equation}
    f(u)= \frac{1}{\gamma} sign \left ( \mid u-v_{th} \mid < \frac{\gamma}{2}\right )
\end{equation} 
Here, $u$ is the neuron membrane potential, $\gamma$ is a hyper-parameter to control the surrogate gradient's width and sharpness, and $v_{th}$ is the threshold.

\section{Experiments and Evaluation} \label{exp}
In the following section, we start by detailing the dataset employed for our evaluations. Next, we outline the hyperparameters used in training SegSNNnet and the metrics chosen to assess our approach. Finally, we analyze and discuss the results obtained.
\subsection{Dataset} \label{dataset_dsec}
The DSEC-semantic dataset, developed by Sun et al. \cite{sun2022ess}, includes $10,891$ RGB frames ($8082$ for the training and $2809$ for the test split) and an event stream between two successive frames. This data was captured across urban and rural environments in Switzerland using automotive-grade standard and high-resolution event cameras. The event stream is represented in the Address Event Representation (AER) format as \((x, y, t, p)\), indicating an event at pixel \((x, y)\) at time \(t\) with polarity \(p \in \{-1, 1\}\). The dataset comprises $11$ classes: background, building, fence, person, pole, road, sidewalk, vegetation, car, wall, and traffic sign. In our experiments, each sample is divided into $3$ regions as it is particularly relevant in the automotive context. Notably, the corners record significantly higher event counts, averaging around $300k$ events, compared to the middle region which captures the scene's perspective and averages $88k$ events. Additionally, we incorporate a $20$-pixel overlap region and implement a reset every $5^{th}$ sample, where data from all regions is forwarded to the SegSNNnet for further processing. 
\subsection{Implementation} \label{impl}
\textbf{Hyperparameters: } We trained the SegSNNnet using a per-pixel CE loss function during our experiments. We employed the Adam optimizer \cite{kingma2014adam} with an initial learning rate of $0.02$, a batch size of $4$, and trained the model for $70$ epochs. The training was conducted on a desktop computer with an NVIDIA RTX A5000 GPU using the PyTorch framework. The SNN network utilized LIF neurons with a hard reset. 

\noindent \textbf{Performance metrics} To evaluate our model and compare it to the baseline, we use $3$ metrics: 
\begin{enumerate}
    \item  Mean Intersection over Union (MIoU) \cite{alonso2019ev}: the standard metric of semantic segmentation :
        \begin{equation} \label{miou}
        \begin{aligned}
              MIoU(y, \widehat y) & = \frac{1}{C} \sum_{j=1}^{C} \frac{\sum_{i=1}^{N} \delta(y_{i,c},1)\delta(y_{i,c}, \widehat{y_{i,c}} )}{\sum_{i=1}^{N} max(1, \delta(y_{i,c},1)\delta(\widehat{y_{i,c}}, 1) )}  \\
              &= \frac{TP}{TP+FP+FN}
         \end{aligned}   
        \end{equation}
        \noindent where, $y$ and $\widehat{y}$ are the ground truth and the network prediction, respectively. $C$ is the number of classes, $N$ is the number of pixels and $\delta $ denotes the Kronecker delta function. TP, TN, FP, and FN stand for true positive, true negative, false positive and false negative, respectively.
    \item Throughput (Number of frames per second = FPS): the larger, the better latency: 
    \begin{equation}
        Throughput = \frac{1}{N_{frames}} \sum_i \frac{1}{segmentation\_i}
    \end{equation}
    \item Number of floating-point operations (flops) \cite{lemaire2022analytical} including ACC and MAC operations measured as followed:
    \begin{equation}
        \#ACC = \theta_{t-1}\times \frac{H_{kernel}}{S} \times \frac{W_{kernel}}{S} \times C_{out} + T\times C_{out} \times H_{out} \times W_{out} +\theta_t
    \end{equation}
    \begin{equation}
        \#MAC =  T\times C_{out} \times H_{out} \times W_{out} 
    \end{equation}
    here, $\theta_{t-1}$ and $\theta_t$ represent the number input and output spikes, respectively. $H_{kernel}$, $W_{kernel}$ and $C_{out}$, $H_{out}$, $ W_{out}$ are the kernel and output dimensions, respectively. $T$ is the number of timesteps, equal to $1$ in our work.
\end{enumerate}

\subsection{Performance results} \label{results} 
\renewcommand{\arraystretch}{1.2} 
\setlength{\tabcolsep}{12pt} 

\begin{table}[bt]
\centering
\caption{Semantic segmentation results on DSEC in terms of MIoU and FPS for various threshold values of the mean number of events. The second row represents the baseline where the threshold is equal to $0$ and the entire image is processed, while the third row involves splitting the input image into $3$ regions and sending all of them to SegSNNnet. The best trade-off between MIoU and FPS is highlighted in bold.}
\begin{tabular}{|c|c|c|}
\hline
\hspace{0pt}Threshold & \hspace{0pt}MIoU {[}\%{]} & \hspace{0pt}FPS \\ \hline
Baseline                  & $54.46 $                           & $151.25$                 \\ \hline
$0$ w/ split                   & $54.19$ $(-0.30)$                  & $47.34$ $(\times 0.27)$  \\ \hline
$4.00$                         & $53.05$ $(-1.41)$                  & $542.62$ $(\times3.57)$ \\ \hline
$4.75$                         & $52.71$ $(-1.75)$                  & $666.52$ $(\times4.47)$  \\ \hline
$5.25$                        & $\textbf{52.52}$ $\textbf{(-1.95)}$                  & $\textbf{759.85}$ $\textbf{(}\boldsymbol{\times}\textbf{5.07)}$  \\ \hline
$6.00$                          & $52.19$ $(-2.27)$                 & $916.82$ $(\times6.08)$  \\ \hline
$8.00$                          & $51.56$ $(-2.90)$                  & $1261.00$ $(\times8.15)$  \\ \hline
$10.0$                         & $\textbf{51.11}$ $\textbf{(-3.35)}$                  & $\textbf{1513.67}$ $\textbf{(}\boldsymbol{\times}\textbf{10.05)}$   \\ \hline
\end{tabular}

\label{tab:main_result}
\end{table}
\begin{table}[tb]
\centering
\caption{Number of flops required to evaluate the entire test set using the baseline (standard approach) and our approach with two different thresholds. Values in brackets indicate the ratio of the baseline value to the corresponding value.}
\label{nb_flops}
\begin{tabular}{|c|c|c|c|}
\hline
Threshold & Baseline                                    & 4.75                               & 10.0                               \\ \hline
\#Flops    & $5,99\times10^{12}$ & $3,30\times10^{12}$ $(\times0.55)$ & $2,17\times10^{12}$ $(\times0.36)$ \\ \hline
\end{tabular}

\end{table}

\setlength{\tabcolsep}{8pt} 
\begin{table}[tb]
\centering
\caption{Comparison of the number of regions processed by SegSNNnet (Column $2$) versus the number of regions obtained through the reuse process (Column $3$).}
\begin{tabular}{|c|ccc|ccc|}
\hline
\multirow{2}{*}{Threshold} & \multicolumn{3}{c|}{\# Regions processed} & \multicolumn{3}{c|}{\# Regions reused} \\ \cline{2-7} 
                                    & Left  & Middle  & Right & Left  & Middle  & Right  \\ \hline
$0$ w/ split                        & 2791           & 2791             & 2791           & 0              & 0                & 0               \\ \hline
$4.00$                              & 1894           & 787              & 1890           & 897            & 2004             & 901             \\ \hline
$4.75$                              & 1711          & 742              & 1747           & 1080           & 2049             & 1086            \\ \hline
$5.25$                              & 1600           & 707              & 1658           & 1191           & 2084             & 1133            \\ \hline
$6.00$                              & 1473           & 680              & 1514           & 1318           & 2111             & 1277            \\ \hline
$8.00$                              & 1200           & 636              & 1266           & 1591           & 2155             & 1525            \\ \hline
$10.0$                              & 1032           & 621              & 1111           & 1759           & 2170             & 1680            \\ \hline
\end{tabular}
\label{tab:nb_regions}
\end{table}

\begin{figure}[!t]
  \centering
  \includegraphics[scale=0.24]{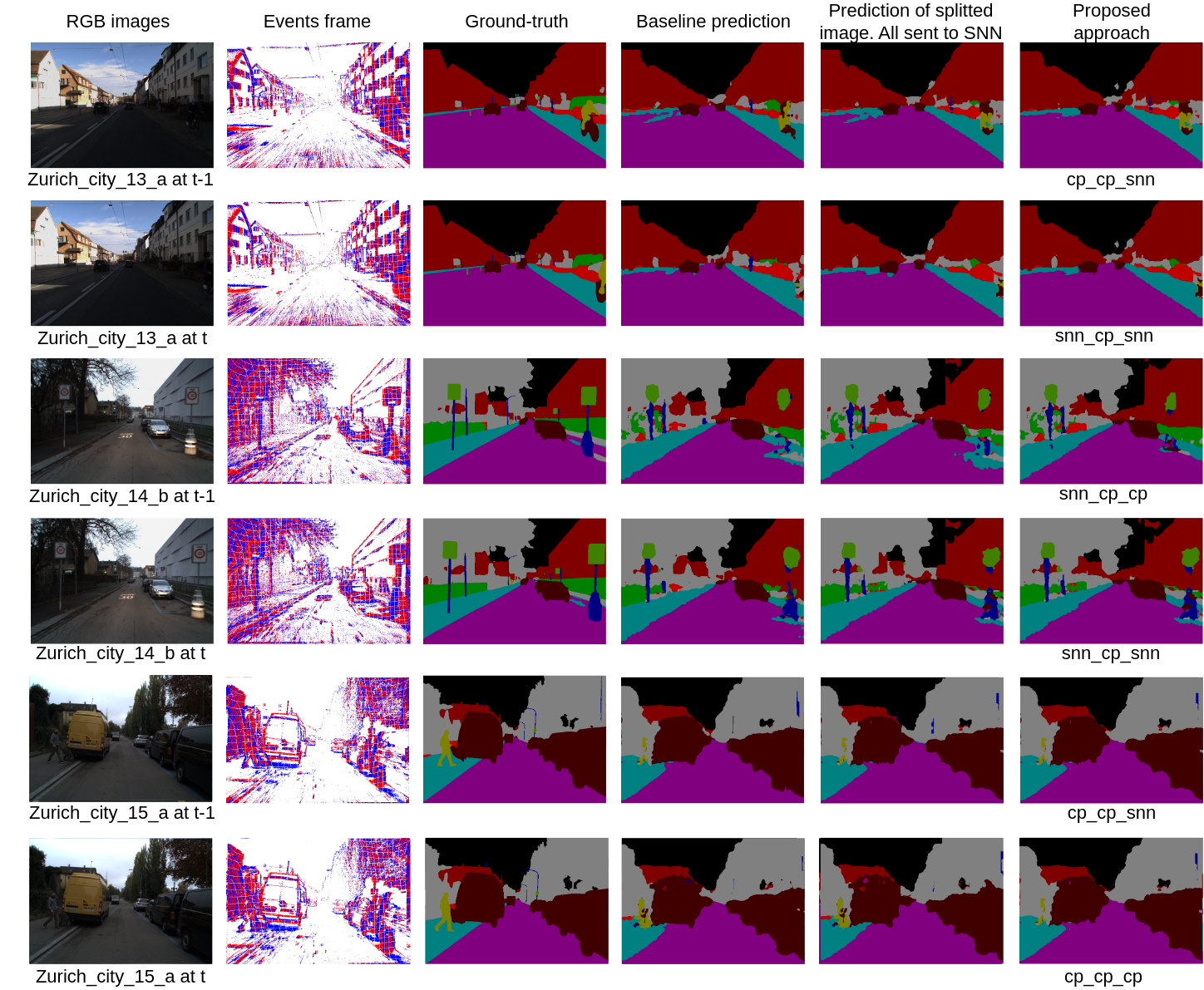}
  \caption{Qualitative results on DSEC Dataset. From left to right, we visualize the RGB images, accumulated events over a $100ms$ interval before the RGB image was captured, ground truth, predictions using the baseline approach, predictions after dividing the image into $3$ regions and processing each with SegSNNnet, and predictions using our method. "CP" indicates the reuse of the keyframe's region by copy-pasting, and "SNN" refers to the processing of the current frame's region by SegSNNnet. }
  \label{fig:qualitative_results}
\end{figure}

\setlength{\tabcolsep}{7pt} 
\begin{table}[!b]
\centering
\caption{Scene-wise MIoU and FPS with varying thresholds. The best trade-off between MIoU and FPS is highlighted in bold.}

\begin{tabular}{|c|cc|cc|cc|}
\hline
\multirow{2}{*}{Threshold} & \multicolumn{2}{c|}{$zurich\_city\_14\_c$}               & \multicolumn{2}{c|}{$zurich\_city\_13\_a$}              & \multicolumn{2}{c|}{$zurich\_city\_15\_a$}              \\ \cline{2-7} 
                           & \multicolumn{1}{c|}{MIoU[\%]}       & FPS              & \multicolumn{1}{c|}{MIoU[\%]}       & FPS             & \multicolumn{1}{c|}{MIoU[\%]}       & FPS             \\ \hline
Baseline                   & \multicolumn{1}{c|}{50.27}          & 151.40           & \multicolumn{1}{c|}{52.62}          & 150.59          & \multicolumn{1}{c|}{54.37}          & 151.34          \\ \hline
$0$ w/ split               & \multicolumn{1}{c|}{50.14}          & 46.21            & \multicolumn{1}{c|}{51.93}          & 46.35           & \multicolumn{1}{c|}{54.16}          & 46.70           \\ \hline
4.00                       & \multicolumn{1}{c|}{48.50}          & 834.55           & \multicolumn{1}{c|}{51.20}          & 636.50          & \multicolumn{1}{c|}{53.49}          & 228.00          \\ \hline
4.75                       & \multicolumn{1}{c|}{\textbf{48.09}} & \textbf{1047.65} & \multicolumn{1}{c|}{50.67}          & 879.96          & \multicolumn{1}{c|}{53.23}          & 332.56          \\ \hline
5.25                       & \multicolumn{1}{c|}{47.84}          & 1133.77          & \multicolumn{1}{c|}{\textbf{50.65}} & \textbf{977.57} & \multicolumn{1}{c|}{53.05}          & 397.23          \\ \hline
6.00                       & \multicolumn{1}{c|}{47.39}          & 1296.16          & \multicolumn{1}{c|}{50.48}          & 1140.99         & \multicolumn{1}{c|}{52.80}          & 538.49          \\ \hline
8.00                       & \multicolumn{1}{c|}{46.67}          & 1659.46          & \multicolumn{1}{c|}{50.14}          & 1396.87         & \multicolumn{1}{c|}{\textbf{52.18}} & \textbf{883.96} \\ \hline
10.0                       & \multicolumn{1}{c|}{46.26}          & 1751.33          & \multicolumn{1}{c|}{49.31}          & 1733.92         & \multicolumn{1}{c|}{51.64}          & 1115.10         \\ \hline
\end{tabular}
\label{tab:results_perscene}
\end{table}
Tab. \ref{tab:main_result} outlines the performance metrics of various semantic segmentation approaches, focusing on MIoU and FPS. In the first row, we compare the proposed approach to a baseline where the standard semantic segmentation process is executed on the entire image using SegSNNnet. In the second, the image is partitioned into $3$ regions, each independently processed by SegSNNnet, aiming to gauge the impact of this division. In the rest, different thresholds for the mean number of events are considered to assess our approach. The results indicate that splitting the image leads to relatively stable performance thanks to the overlapping. It also highlights that our approach substantially reduces latency, achieving a $5.07\times$ or $10.05\times$ increase in FPS while experiencing only a $1.95\%$ or $3.35\%$ decrease in MIoU. Additionally, it is observed that as the threshold increases, the enhancement in FPS outweighs the decline in MIoU. Tab. \ref{tab:nb_regions} explains the underlying trends observed in the results. With increasing thresholds, the number of regions processed by the semantic segmentation network progressively diminishes. Notably, the number of middle regions processed by SegSNNnet is much lower than the other two due to the lower event density in that area. Consequently, as shown in Tab. \ref{nb_flops}, the number of flops required to process the entire test set in the baseline is $5.99\times10^{12}$, which is $1.81\times$ and $2.75\times$ higher than the number of flops required in our approach using the thresholds of $4.75$ and $10$, respectively, thus reducing the energy consumption. Furthermore, a detailed analysis of the dataset fold representing various rural scenes, as detailed in Tab. \ref{tab:results_perscene}, reveals that the optimal trade-off between MIoU and FPS is not consistently achieved with the same threshold value. Hence, this opens a potential avenue for future research to define this threshold across diverse scenes dynamically.
Moreover, Fig. \ref{fig:qualitative_results} showcases the qualitative outcomes of our experiments. These results, obtained at a frame frequency of $10Hz$, allow for a clearer observation of image differences and their respective segmentations compared to $20Hz$ used in our experiments. With a  $\theta=2.1$ and an overlap region of $20$ pixels, the MIoU stands at $51.25\%$, representing a marginal loss of $2.44\%$, with an FPS $4.12\times$ higher than the baseline. Consequently, the qualitative results from our experiments surpass those depicted in the figure, boasting higher MIoU and FPS metrics.
 
\noindent 
Finally, we benchmark our approach against the existing methods for semantic segmentation using DSEC, focusing solely on MIoU, as information regarding FPS in these works is unavailable. Sun et al. \cite{sun2022ess} utilize events represented as a voxel grid in conjunction with a standard ANN model. On the other hand, Das et al. \cite{das2024halsie} employ a hybrid network architecture, where an ANN processes RGB data and an SNN handles event data, with the fused features subsequently fed into another ANN. Tab. \ref{tab:comparison_sota_dsec} illustrates that our approach is competitive with the state-of-the-art methods while employing significantly fewer parameters. Thus, contributing to reduced latency and facilitating easier implementation on embedded hardware.

\begin{table}[tb]
\centering
\caption{Comparison of different methods using events (E) or events and frames (E+F) for semantic segmentation on DSEC-semantic. E$^*$ refers to using the raw event stream directly, without transforming it into another structure for input.}
\label{tab:comparison_sota_dsec}
\begin{tabular}{|c|c|c|c|c|}
\hline
Method & Type & Params [M] & Input &MIoU  [\%] \\
\hline
ESS \cite{sun2022ess}                                & ANN &$12.91$& E                  & $53.29$ \\ 
HALSIE \cite{das2024halsie}                               & Hybrid & $1.82$&E+F              & $52.43$ \\  
Proposed approach                                          & SNN   &$0.33$ & E$^*$+F              & $52.52$ \\ \hline 

\end{tabular}
\end{table}
\subsection{Ablation study} \label{ablation}
After thoroughly analyzing the results of our approach, we proceed with an ablation study to further dissect its components. Initially, we explore replacing the reuse process of the semantic segmentation result with optical flow, inspired by the Xu et al. approach \cite{xu2018dynamic}. Then, we investigate varying the number of regions processed by our approach to assess its impact on performance metrics. This step aims to elucidate how the segmentation and processing strategy of different region configurations influence overall effectiveness.
\\
\noindent\textbf{Optical flow:}
Building upon the methodology introduced by Xu et al. \cite{xu2018dynamic}, we aimed to augment the outcomes by integrating a warp function instead of straightforwardly reusing the semantic segmentation result of the keyframe to get the current frame segmentation result. This warp function leverages bilinear interpolation on the keyframe's semantic segmentation whose pixels are shifted according to the displacement of the pixels represented on the optical flow generated between successive frames. For these experiments, instead of using \textit{'$zurich\_city\_14\_c$'}, \textit{'$zurich_city\_13\_a$'}, and \textit{'$zurich\_city\_15\_a$'} folds, we used \textit{'$zurich\_city\_01\_a$'}, \textit{'$zurich\_city\_02\_a$'}, and \textit{'$zurich_city\_08\_a$'} with a frame frequency of $10Hz$, as optical flow ground truth is available only for these samples. The results are summarized in Tab \ref{of_results}, revealing that the optical flow inclusion marginally enhances segmentation performance in terms of MIoU. However, it notably increases latency, transitioning from $155$ Fps in the baseline to $5.40$ Fps. This can be attributed to the higher number of parameters in the optical flow, which is twice as many as those in SegSNNnet, along with the additional processing required for the warp function.

\setlength{\tabcolsep}{4pt} 

\begin{table}[t]
\centering
\caption{Comparison of semantic segmentation results on DSEC using MIoU and FPS metrics at various mean event thresholds for our approach (Up for update) versus optical flow (OF).}
\label{of_results}
\begin{tabular}{|c|c|c|c|c|}
\hline
\hspace{0pt}\textbf{Threshold} &  \hspace{0pt}\textbf{MIoU$\_$OF {[}\%{]}} &\hspace{0pt}\textbf{FPS$\_$OF} & \hspace{0pt}\textbf{MIoU$\_$Up {[}\%{]}} & \hspace{0pt}\textbf{FPS$\_$Up} \\ \hline
Baseline                       & $ 52.60$                                            & $155.82$                          & $ 52.60$                                            & $155.82$                          \\ \hline
$0$ w/ split                        & $52.22$ $(-0.38)$                                   & $5.41$ $(\times0.03)$          & $52.22$ $(-0.38)$                                   & $58.97$ $(\times0.37)$          \\ \hline
$3.50$                              & $51.47$ $(-1.14)$                                   & $5.42$ $(\times0.03)$          & $50.92$ $(-1.68)$                                   & $180.58$ $(\times1.16)$           \\ \hline
$4.00$                              & $51.46$ $(-1.14)$                                   & $5.44$ $(\times0.03)$          & $50.74$ $(-1.86)$                                   & $206.50$ $(\times1.33)$           \\ \hline
$4.50$                              & $51.19$ $(-1.42)$                                   & $5.45$ $(\times0.03)$          & $50.31$ $(-2.29)$                                   & $262.47$ $(\times1.68)$           \\ \hline
$5.00$                              & $50.91$ $(-1.70)$                                   & $5.42$ $(\times0.03$          & $49.92$ $(-2.68)$                                   & $416.15$ $(\times2.67)$           \\ \hline
$6.00$                              & $50.76$ $(-1.84)$                                   & $5.44$ $(\times0.03)$          & $49.46$ $(-3.15)$                                   & $533.90$ $(\times3.20)$           \\ \hline
\end{tabular}
\end{table}

\noindent The variation in results between our study and the original paper stems from differences in model parameters. Their semantic segmentation model has $38M$ parameters, a depth that is $12\times$ greater than the $3M$ parameters in their optical flow network. Consequently, their approach achieves faster processing by utilizing the optical flow network. In contrast, as mentioned above, our model comprises $0.33M$ parameters, multiplying by $4$ the parameter count of the Cuadrado et al. model \cite{cuadrado2023optical} which contains $1.2M$ parameters. 
\noindent It is noteworthy that the results obtained using our approach in this instance may be less compelling compared to those presented in Tab \ref{tab:main_result}. This discrepancy can be attributed to $2$ factors: (1) The limited number of samples used to train our model, which stands at $3177$ instead of the original $8082$ training samples. (2) The lower frame frequency employed in this scenario which contributes to the reduced similarity between consecutive frames.

\noindent\textbf{Number of regions: }
After visualizing the DSEC dataset, we observed that the number of events is significantly higher in the corners of the images compared to the middle. Consequently, we initially divided the input images into $3$ regions, as shown in Fig. \ref{fig:overview}. In this section, however, we explore the impact of splitting the image into $3$ regions but horizontally, $4$ regions (2x2) and $9$ regions (3x3). The results, summarized in Tab. \ref{tab:vary_regions}, demonstrate that our initial three-region division is more effective. Increasing the number of regions leads to a decrease in FPS while maintaining almost the same MIoU for $4$ regions and decreasing it for $3$ horizontal and $9$ regions. This decline is due to SegSNNnet being trained on $640\times440$ images, whereas the inference for $3$ horizontal and $9$ regions is conducted on different dimensions of $640\times166$ and $233\times166$, respectively.
\begin{table}[tb]
\centering
\caption{Comparison of semantic segmentation results on DSEC in terms of MIoU and FPS, while varying the number of regions into which the input image is split. }
\begin{tabular}{|c|c|c|c|}
\hline
\# Regions          & Threshold & MIoU [\%]     & FPS                   \\ \hline
\multirow{2}{*}{3 $(1\times3$)} & $0.00$     & $54.19$         & $47.34$                 \\ \cline{2-4} 
                   & $5.25$     & $52.52$ $(-1.95)$ & $759.85$ $(\times5.07)$  \\ \hline
\multirow{2}{*}{3 $(3\times1$)} & $0.00$     & $50.92$         & $24.31$                 \\ \cline{2-4} 
                   & $3.50$     & $50.21$ $(-4.25)$ & $521.16$ $(\times3.47)$  \\ \hline                   
\multirow{2}{*}{4 $(2\times2)$} & $0.00$     & $52.99$         & $43.14$                 \\ \cline{2-4} 
                   & $6.00$     & $50.81$ $(-3.65)$ & $500.75$ $(\times3.37)$  \\ \hline
\multirow{2}{*}{9 $(3\times3)$} & $0.00$     & $50.60$     &          $19.38$             \\ \cline{2-4} 
                   & $4.00$     & $48.18$ $(-6.28)$ & $65.52$ $(\times0.43)$ \\ \hline
\end{tabular}

\label{tab:vary_regions}
\end{table}



\section{Conclusion} \label{sec_conclusion}
This paper presents a novel approach for neuromorphic semantic segmentation that leverages event streams to achieve a balanced trade-off between reduced latency, minimal accuracy loss, and energy efficiency. Our evaluation on the DSEC dataset demonstrates a throughput increase of up to $5\times$ and $10\times$ over the baseline, with only a minimal reduction in MIoU of $1.94\%$ and $3.35\%$, respectively. Additionally, our method significantly reduces energy consumption by decreasing the number of flops and using a spiking neural network, activating just $19.20\%$ of neurons during inference. The small number of parameters and the generation of binary matrices or integers in our model facilitate easy deployment on SPLEAT, a neuromorphic architecture designed for SNNs on ASIC and FPGA hardware. Future work will focus on refining this method by introducing dynamic thresholds and adaptive input image splitting. We also aim to address the current limitation of detecting new objects in frames with a low number of events, ensuring our approach remains robust and versatile across various scenarios.

\section*{Acknowledgements} 
\noindent This work was supported by the project NAMED (ANR-23-CE45-0025-01) of the French National Research Agency (ANR).
\\
\noindent This work was supported by the French government through the France 2030 investment plan managed by the National Research Agency (ANR), as part of the Initiative of Excellence Université Côte d'Azur under reference number ANR-15-IDEX-01.
\\
\noindent This material is based upon work supported by the French national research agency (project DEEPSEE ANR-17-CE24-0036)
\\
\noindent The authors are grateful to the OPAL infrastructure from Université Côte d'Azur for providing resources and support.

%
%
%
\bibliographystyle{splncs04}
\bibliography{biblio}

\begin{thebibliography}{10}
\providecommand{\url}[1]{\texttt{#1}}
\providecommand{\urlprefix}{URL }
\providecommand{\doi}[1]{https://doi.org/#1}

\bibitem{abderrahmane2022spleat}
Abderrahmane, N., Miramond, B., Kervennic, E., Girard, A.: Spleat: Spiking low-power event-based architecture for in-orbit processing of satellite imagery. In: 2022 International Joint Conference on Neural Networks (IJCNN). pp. 1--10. IEEE (2022)

\bibitem{alonso2019ev}
Alonso, I., Murillo, A.C.: Ev-segnet: Semantic segmentation for event-based cameras. In: Proceedings of the IEEE/CVF Conference on Computer Vision and Pattern Recognition Workshops. pp.~0--0 (2019)

\bibitem{cordts2016cityscapes}
Cordts, M., Omran, M., Ramos, S., Rehfeld, T., Enzweiler, M., Benenson, R., Franke, U., Roth, S., Schiele, B.: The cityscapes dataset for semantic urban scene understanding. In: Proceedings of the IEEE conference on computer vision and pattern recognition. pp. 3213--3223 (2016)

\bibitem{cuadrado2023optical}
Cuadrado, J., Ran{\c{c}}on, U., Cottereau, B.R., Barranco, F., Masquelier, T.: Optical flow estimation from event-based cameras and spiking neural networks. Frontiers in Neuroscience  \textbf{17},  1160034 (2023)

\bibitem{dampfhoffer2023leveraging}
Dampfhoffer, M., Mesquida, T., Hardy, E., Valentian, A., Anghel, L.: Leveraging sparsity with spiking recurrent neural networks for energy-efficient keyword spotting. In: ICASSP 2023-2023 IEEE International Conference on Acoustics, Speech and Signal Processing (ICASSP). pp.~1--5. IEEE (2023)

\bibitem{das2024halsie}
Das~Biswas, S., Kosta, A., Liyanagedera, C., Apolinario, M., Roy, K.: Halsie: Hybrid approach to learning segmentation by simultaneously exploiting image and event modalities. In: Proceedings of the IEEE/CVF Winter Conference on Applications of Computer Vision. pp. 5964--5974 (2024)

\bibitem{davies2018loihi}
Davies, M., Srinivasa, N., Lin, T.H., Chinya, G., Cao, Y., Choday, S.H., Dimou, G., Joshi, P., Imam, N., Jain, S., et~al.: Loihi: A neuromorphic manycore processor with on-chip learning. Ieee Micro  \textbf{38}(1),  82--99 (2018)

\bibitem{fang2021deep}
Fang, W., Yu, Z., Chen, Y., Huang, T., Masquelier, T., Tian, Y.: Deep residual learning in spiking neural networks. Advances in Neural Information Processing Systems  \textbf{34},  21056--21069 (2021)

\bibitem{furber2014spinnaker}
Furber, S.B., Galluppi, F., Temple, S., Plana, L.A.: The spinnaker project. Proceedings of the IEEE  \textbf{102}(5),  652--665 (2014)

\bibitem{gehrig2020video}
Gehrig, D., Gehrig, M., Hidalgo-Carri{\'o}, J., Scaramuzza, D.: Video to events: Recycling video datasets for event cameras. In: Proceedings of the IEEE/CVF Conference on Computer Vision and Pattern Recognition. pp. 3586--3595 (2020)

\bibitem{gehrig2019end}
Gehrig, D., Loquercio, A., Derpanis, K.G., Scaramuzza, D.: End-to-end learning of representations for asynchronous event-based data. In: Proceedings of the IEEE/CVF International Conference on Computer Vision. pp. 5633--5643 (2019)

\bibitem{guo2023joint}
Guo, Y., Peng, W., Chen, Y., Zhang, L., Liu, X., Huang, X., Ma, Z.: Joint a-snn: Joint training of artificial and spiking neural networks via self-distillation and weight factorization. Pattern Recognition  \textbf{142},  109639 (2023)

\bibitem{hareb2024evsegsnn}
Hareb, D., Martinet, J.: Evsegsnn: Neuromorphic semantic segmentation for event data. arXiv preprint arXiv:2406.14178  (2024)

\bibitem{jain2019accel}
Jain, S., Wang, X., Gonzalez, J.E.: Accel: A corrective fusion network for efficient semantic segmentation on video. In: Proceedings of the IEEE/CVF Conference on Computer Vision and Pattern Recognition. pp. 8866--8875 (2019)

\bibitem{khacef2018confronting}
Khacef, L., Abderrahmane, N., Miramond, B.: Confronting machine-learning with neuroscience for neuromorphic architectures design. In: 2018 International Joint Conference on Neural Networks (IJCNN). pp.~1--8. IEEE (2018)

\bibitem{kim2022beyond}
Kim, Y., Chough, J., Panda, P.: Beyond classification: Directly training spiking neural networks for semantic segmentation. Neuromorphic Computing and Engineering  \textbf{2}(4),  044015 (2022)

\bibitem{kingma2014adam}
Kingma, D.P., Ba, J.: Adam: A method for stochastic optimization. arXiv preprint arXiv:1412.6980  (2014)

\bibitem{lemaire2022analytical}
Lemaire, E., Cordone, L., Castagnetti, A., Novac, P.E., Courtois, J., Miramond, B.: An analytical estimation of spiking neural networks energy efficiency. In: International Conference on Neural Information Processing. pp. 574--587. Springer (2022)

\bibitem{Spleat_application2020}
Lemaire, E., Moretti, M., Daniel, L., Miramond, B., Millet, P., Feresin, F., Bilavarn, S.: An fpga-based hybrid neural network accelerator for embedded satellite image classification. In: 2020 IEEE International Symposium on Circuits and Systems (ISCAS) (2020). \doi{10.1109/ISCAS45731.2020.9180625}

\bibitem{li2018low}
Li, Y., Shi, J., Lin, D.: Low-latency video semantic segmentation. In: Proceedings of the IEEE conference on Computer Vision and Pattern Recognition. pp. 5997--6005 (2018)

\bibitem{maass1997networks}
Maass, W.: Networks of spiking neurons: the third generation of neural network models. Neural networks  \textbf{10}(9),  1659--1671 (1997)

\bibitem{nahmias2013leaky}
Nahmias, M.A., Shastri, B.J., Tait, A.N., Prucnal, P.R.: A leaky integrate-and-fire laser neuron for ultrafast cognitive computing. IEEE journal of selected topics in quantum electronics  \textbf{19}(5),  1--12 (2013)

\bibitem{neftci2019surrogate}
Neftci, E.O., Mostafa, H., Zenke, F.: Surrogate gradient learning in spiking neural networks: Bringing the power of gradient-based optimization to spiking neural networks. IEEE Signal Processing Magazine  \textbf{36}(6),  51--63 (2019)

\bibitem{sun2022ess}
Sun, Z., Messikommer, N., Gehrig, D., Scaramuzza, D.: Ess: Learning event-based semantic segmentation from still images. In: European Conference on Computer Vision. pp. 341--357. Springer (2022)

\bibitem{wang2021dual}
Wang, L., Chae, Y., Yoon, K.J.: Dual transfer learning for event-based end-task prediction via pluggable event to image translation. In: Proceedings of the IEEE/CVF International Conference on Computer Vision. pp. 2135--2145 (2021)

\bibitem{wang2021evdistill}
Wang, L., Chae, Y., Yoon, S.H., Kim, T.K., Yoon, K.J.: Evdistill: Asynchronous events to end-task learning via bidirectional reconstruction-guided cross-modal knowledge distillation. In: Proceedings of the IEEE/CVF Conference on Computer Vision and Pattern Recognition. pp. 608--619 (2021)

\bibitem{wang2021end}
Wang, Y., Xu, Z., Wang, X., Shen, C., Cheng, B., Shen, H., Xia, H.: End-to-end video instance segmentation with transformers. In: Proceedings of the IEEE/CVF conference on computer vision and pattern recognition. pp. 8741--8750 (2021)

\bibitem{wu2022seqformer}
Wu, J., Jiang, Y., Bai, S., Zhang, W., Bai, X.: Seqformer: Sequential transformer for video instance segmentation. In: European Conference on Computer Vision. pp. 553--569. Springer (2022)

\bibitem{wu2019direct}
Wu, Y., Deng, L., Li, G., Zhu, J., Xie, Y., Shi, L.: Direct training for spiking neural networks: Faster, larger, better. In: Proceedings of the AAAI conference on artificial intelligence. vol.~33, pp. 1311--1318 (2019)

\bibitem{xu2018dynamic}
Xu, Y.S., Fu, T.J., Yang, H.K., Lee, C.Y.: Dynamic video segmentation network. In: Proceedings of the IEEE conference on computer vision and pattern recognition. pp. 6556--6565 (2018)

\bibitem{yang2022temporally}
Yang, S., Wang, X., Li, Y., Fang, Y., Fang, J., Liu, W., Zhao, X., Shan, Y.: Temporally efficient vision transformer for video instance segmentation. In: Proceedings of the IEEE/CVF conference on computer vision and pattern recognition. pp. 2885--2895 (2022)

\bibitem{zhang2023energy}
Zhang, H., Fan, X., Zhang, Y.: Energy-efficient spiking segmenter for frame and event-based images. Biomimetics  \textbf{8}(4), ~356 (2023)

\bibitem{zhu2017deep}
Zhu, X., Xiong, Y., Dai, J., Yuan, L., Wei, Y.: Deep feature flow for video recognition. In: Proceedings of the IEEE conference on computer vision and pattern recognition. pp. 2349--2358 (2017)

\end{thebibliography}

\end{document}